\documentclass[letterpaper, 10pt, conference]{ieeeconf}

\usepackage[left=54pt,top=50pt,right=54pt,bottom=57pt]{geometry}

\usepackage{times}
\usepackage{tabularx}
\usepackage{subcaption}
\usepackage{multirow,multicol, array}
\usepackage[font={small}]{caption}   
\usepackage{graphicx,dblfloatfix}
\usepackage{wrapfig}
\usepackage{siunitx}

\let\labelindent\relax
\usepackage{enumitem}

\usepackage{amsmath,amsthm,amssymb,amsfonts}
\usepackage[titlenumbered,ruled]{algorithm2e}
\usepackage{textcomp}
\usepackage{acronym}
\usepackage{balance}
\usepackage{mdwmath}
\usepackage{bm}
\usepackage{mdwtab}
\usepackage{array}
\usepackage{eqparbox}
\usepackage{cite}

\usepackage{color}
\usepackage{psfrag}
\usepackage{epsfig}
\usepackage{url}
\usepackage{epstopdf}
\usepackage{booktabs}
\usepackage{blindtext}

\usepackage[colorlinks,bookmarksopen,bookmarksnumbered,linkcolor=blue,citecolor=blue,urlcolor=blue]{hyperref}

\usepackage[colorinlistoftodos,prependcaption,textsize=tiny]{todonotes}

\usepackage[a-1b]{pdfx}

\renewcommand{\emph}{\textit}

\newtheorem*{lemma*}{Lemma}

\newtheorem*{problem*}{Problem}

\newcommand{\sgn}{\text{sign}}

\makeatletter
\newcommand\fs@spaceruled{\def\@fs@cfont{\bfseries}\let\@fs@capt\floatc@ruled
    \def\@fs@pre{\vspace{5\baselineskip}\hrule height.8pt depth0pt \kern2pt}%
    \def\@fs@post{\kern2pt\hrule\relax}%
    \def\@fs@mid{\kern2pt\hrule\kern2pt}%
    \let\@fs@iftopcapt\iftrue}
\makeatother

\IEEEoverridecommandlockouts
\graphicspath{{images/}}

\begin{document}

	
	\title{Deformation Recovery Control and Post-Impact Trajectory Replanning \\for Collision-Resilient Mobile Robots}
	
	\author{Zhouyu Lu, Zhichao Liu, and Konstantinos Karydis
		\thanks{The authors are with the Dept. of Electrical and Computer Engineering, University of California, Riverside. 
			Email: {\{zlu044, zliu157, karydis\}@ucr.edu}. 
		We gratefully acknowledge the support of NSF 
		\#IIS-1910087, ONR 
		\#N00014-18-1-2252 and \#N00014-19-1-2264, and ARL 
		\#W911NF-18-1-0266. 
		Any opinions, findings, and conclusions or recommendations expressed in this material are those of the authors and do not necessarily reflect the views of the funding agencies.}
	}

	\maketitle
	\thispagestyle{empty}

\begin{abstract}
The paper focuses on collision-inclusive motion planning for impact-resilient mobile robots. We propose a new deformation recovery and replanning strategy to handle collisions that may occur at run-time. Contrary to collision avoidance methods that generate trajectories only in conservative local space or require collision checking that has high computational cost, our method directly generates (local) trajectories with imposing only waypoint constraints. If a collision occurs, our method then estimates the post-impact state and computes from there an intermediate waypoint to recover from the collision. To achieve so, we develop two novel components: 1) a deformation recovery controller that optimizes the robot's states during post-impact recovery phase, and 2) a post-impact trajectory replanner that adjusts the next waypoint with the information from the collision for the robot to pass through and generates a polynomial-based minimum effort trajectory. The proposed strategy is evaluated experimentally with an omni-directional impact-resilient wheeled robot. The robot is designed in house, and it can perceive collisions with the aid of Hall effect sensors embodied between the robot's main chassis and a surrounding deflection ring-like structure.
%
%
	%
\end{abstract}
\vspace{-0pt}
	
\section{Introduction}
Mobile robot motion planning algorithms can be classified in terms of the underlying optimization problem~\cite{tordesillas2019faster}. When there exist obstacles, most algorithms generate collision-free trajectories. Yet, it may be possible that potential collisions can in fact be useful~\cite{schmickl2009get,karydis2014planning,haldane2016robotic,mulgaonkar2017robust,mayya2018localization,StagerISER18,khedekar2019contact,mulgaonkar2020tiercel}, and thus an increasing number of research efforts aims at developing collision-inclusive motion planning algorithms~\cite{lu2019optimal,lu2020motion,mote2020collision, zha2020exploiting}. 
This paper focuses on the latter category of collision-inclusive motion planning.

\begin{figure}[!t]
\vspace{6pt}
\centering
\includegraphics[trim={3cm 2.85cm 3cm 0.85cm}, clip, width=0.3\textwidth]{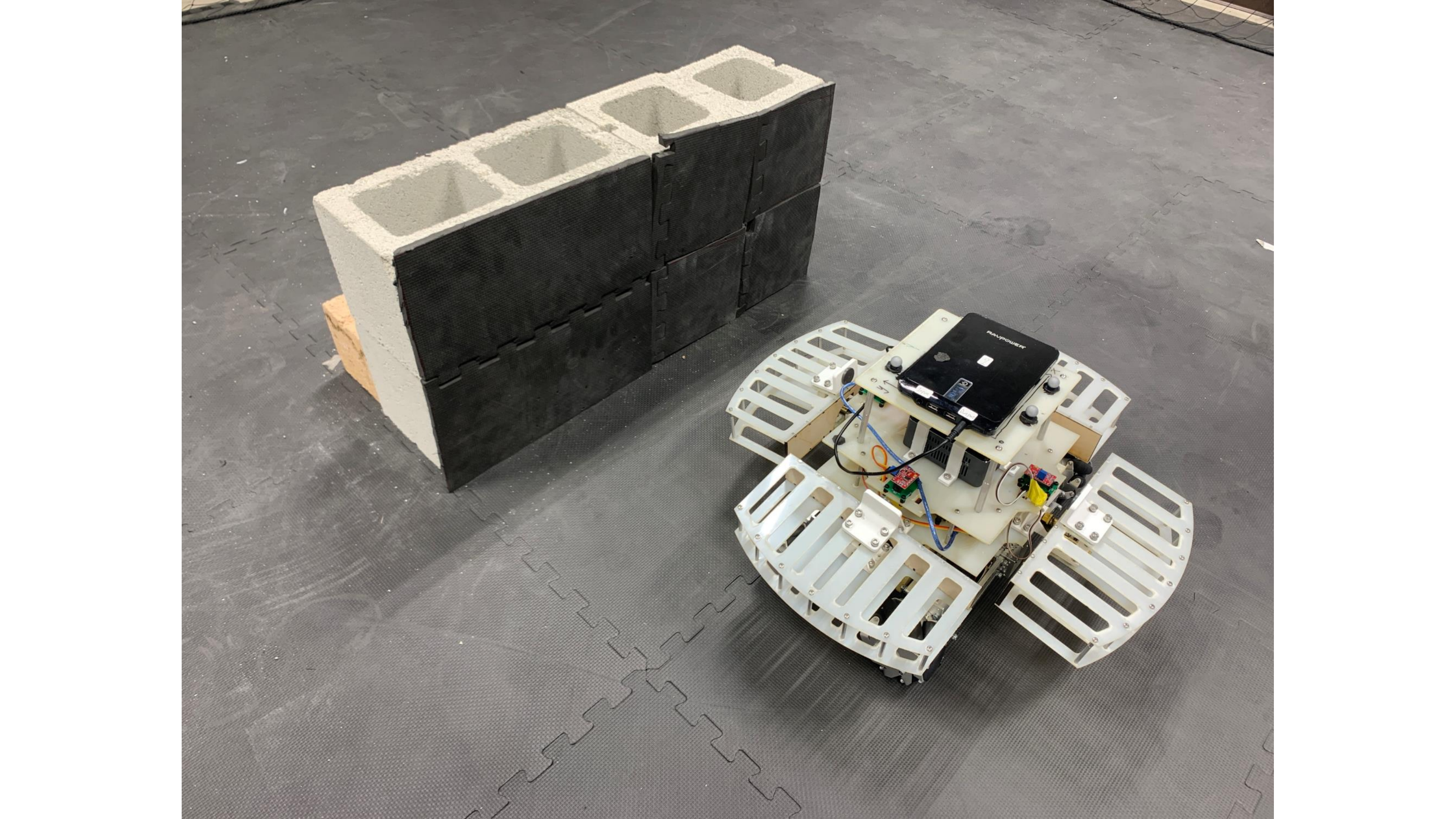}
\vspace{-1pt}
\caption{Snapshot of our built-in-house omni-directional robot. 
The supplementary video contains detailed instances of our experiments. 
      }
      \label{fig:introFIG}
	\vspace{-15pt}
\end{figure}

Development of collision-inclusive motion planners requires three core capabilities: 1) collision resilience, 2) collision identification, and 3) post-impact characterization. Advances in material science and design have helped introduce a range of collision-resilient mobile robots (e.g.,~\cite{briod2014collision,haldane2015integrated,stager2016stochastic,li2019agile}). For instance, passive protection devices can protect the robot from catastrophic impacts but cannot acquire information on where a collision occurred~\cite{StagerISER18}; this is important information in order to predict how the robot will respond following a collision~\cite{StagerICRA19, mulgaonkar2017robust}. 
Sensor-based collision detection and characterization methods have mostly focused on utilizing data from an onboard inertial measurement unit (IMU)~\cite{battiston2019attitude}. However, IMUs are usually unable to distinguish collisions during aggressive maneuvers and to detect static contacts, resulting in low accuracy in collision detection.  Hall effect sensors have been used in the past to provide more accurate collision detection based on estimated deformation \cite{briod2013contact}. In related yet distinct previous work~\cite{liu2021toward}, we have implemented a passive quadrotor arm design with Hall effect sensors, making the robot able to detect and characterize collisions. Herein, we embed a similar design to connect the main robot chassis and the deflection ring to estimate and characterize collisions. We design and deploy a distributed system of collision guards (that form a deflection `ring' as shown in  Fig.~\ref{fig:introFIG}), where each surface is connected to the main robot chassis via passive visco-elastic prismatic joints. 

Contrary to collision avoidance methods that generate trajectories only in conservative local space or require collision checking that has high computational cost and accurate detection of the obstacle, our method directly generates local trajectories based on a minimal effort optimization with only a waypoint constraint. Our method then estimates the post-impact state if a collision occurs and computes an intermediate trajectory to recover from the collision. The refined robot trajectory after collision attains a piece-wise polynomial form with respect to its flat outputs, and is computed online. 
%
%
%
Succinctly, in this work:
%
	\begin{itemize}
		\item We propose and solve a deformation control problem based on a dynamic model of wheeled robot visco-elastic collisions with polygon-shaped obstacles. 
		\item We present a post-impact trajectory replanning algorithm based on constrained quadratic programming (QP) to locally adjust the trajectory after a collision.
		\item We develop and evaluate experimentally a recovery deformation and replanning (DRR) strategy to generate local trajectories based on any given waypoints.
	\end{itemize}
    
    
\section{Related Works}
Most motion planning algorithms focus on generating collision-free trajectories. Collision-free algorithms handle obstacle avoidance in distinct ways. One way is to include geometric constraints of collision-free trajectories within the optimization problem, when a map of the environment is exactly known~\cite{preiss2017trajectory}. When the map is partially-known, it is possible to formulate an obstacle-agnostic optimization problem~\cite{liu2017search, zhou2020robust}. However, this approach is limited to short look-ahead planned trajectories, and may be unable to perform complex maneuvers around obstacles; when they do, this usually requires a computationally-expensive search to be able to generate a trajectory around obstacles. 
	Another approach is to include the obstacles directly in the optimization problem, such as in the form of safe corridor constraints~\cite{liu2017planning}. This way relies on decomposing the free (known or sensed) space as a series of $P$ overlapping polyhedra. However, trajectories generated based on convex decomposition methods can be conservative since the solver can only choose to place the two extreme points of each interval in the overlapping area of two consecutive polyhedra. To overcome conservative solutions, it is possible to use binary variables to allow the solver to choose the specific interval allocation~\cite{deits2015efficient}, but with increased computation time. 

Importantly, as robots venture into more dynamic, irregularly-shaped and obstacle-cluttered environments, avoiding collisions becomes a major challenge~\cite{hoy2015algorithms, campbell2012review}. Employing a conservative local collision avoidance planner in cluttered environments may preclude the robot from finding a feasible path to the goal even if such path exists~\cite{oleynikova2018safe}. Further, detecting all obstacles in the environment can also be a challenging task, especially for those obstacles that are not opaque~\cite{mulgaonkar2020tiercel}, such as glass or reflective surfaces. Developing collision-inclusive motion planners can help address the aforementioned challenges. 


Post-impact characterization in collision-inclusive motion planning can be challenging to achieve, and it directly affects the form that post-impact (recovery) trajectories can attain. 
Estimating an accurate state of the robot after the collision is important for generating a reliable recovery trajectory. 
Yet, modeling collisions is a challenge and requires consideration of many interacting physical phenomena relating to the geometric, material, and inertial properties of each body involved in the collision; many of these properties are themselves difficult to model accurately. 
Related work~\cite{mote2020collision} has introduced an empirical algebraic collision model, by directly relating pre- and post-impact velocities with no thrust commanded. This approach can only deal with a specific pair of objects over a relatively limited range of conditions. 
Other works redirecting the robot after a collision rely on the impulse contact model and feature different stages to track the recovery trajectory~\cite{dicker2018recovery}. An impulse contact model assumes that velocity changes happen instantaneously, thus resulting in discontinuities in the recovery trajectory. 
The robot trajectory after contact with obstacles could be smoother if passive shock absorption devices were to be employed, e.g., as in~\cite{senoo2016deformation,TanakaICRA20} albeit in a different context of robotic arms.
%
%
Herein we develop a dynamic model of our wheeled robot after (visco-elastic) collisions based on the Voigt model for passive arms~\cite{senoo2016deformation}.

Compared to our previous work~\cite{lu2020motion} that developed a higher-level waypoint planner that trades-off between risk and collision exploitation, this paper proposes a new lower-level planner to generate local trajectories based on given waypoints and modify them accordingly if collisions occur.  
%

Our proposed method shares similarities with blind navigation techniques (e.g.,~\cite{buniyamin2011simple,briod2013contact}) in that it requires limited sensing capabilities (only mechanoception via the passive compliant deflectors), and that it can be tuned to recover paths yielded by such methods as well. However, our proposed approach demonstrates features that distinct it from prior blind navigation methods; namely, its ability to exploit collisions in a controlled manner to improve overall collision-inclusive planning performance, its ability to be tuned to switch between exploration (i.e. explore free space after collision) and exploitation (i.e. follow obstacle surface after collision), and its compatibility to work alongside a range of other motion planners (it requires a list of waypoints that can be produced in any means).


\section{The Deformation Recovery and \\ Replanning (DRR) Strategy}
\begin{figure}[!h]
\vspace{-12pt}
\centering
\includegraphics[trim=0.5cm 0.1cm 1.0cm 0.5cm, clip,width=0.825\linewidth]{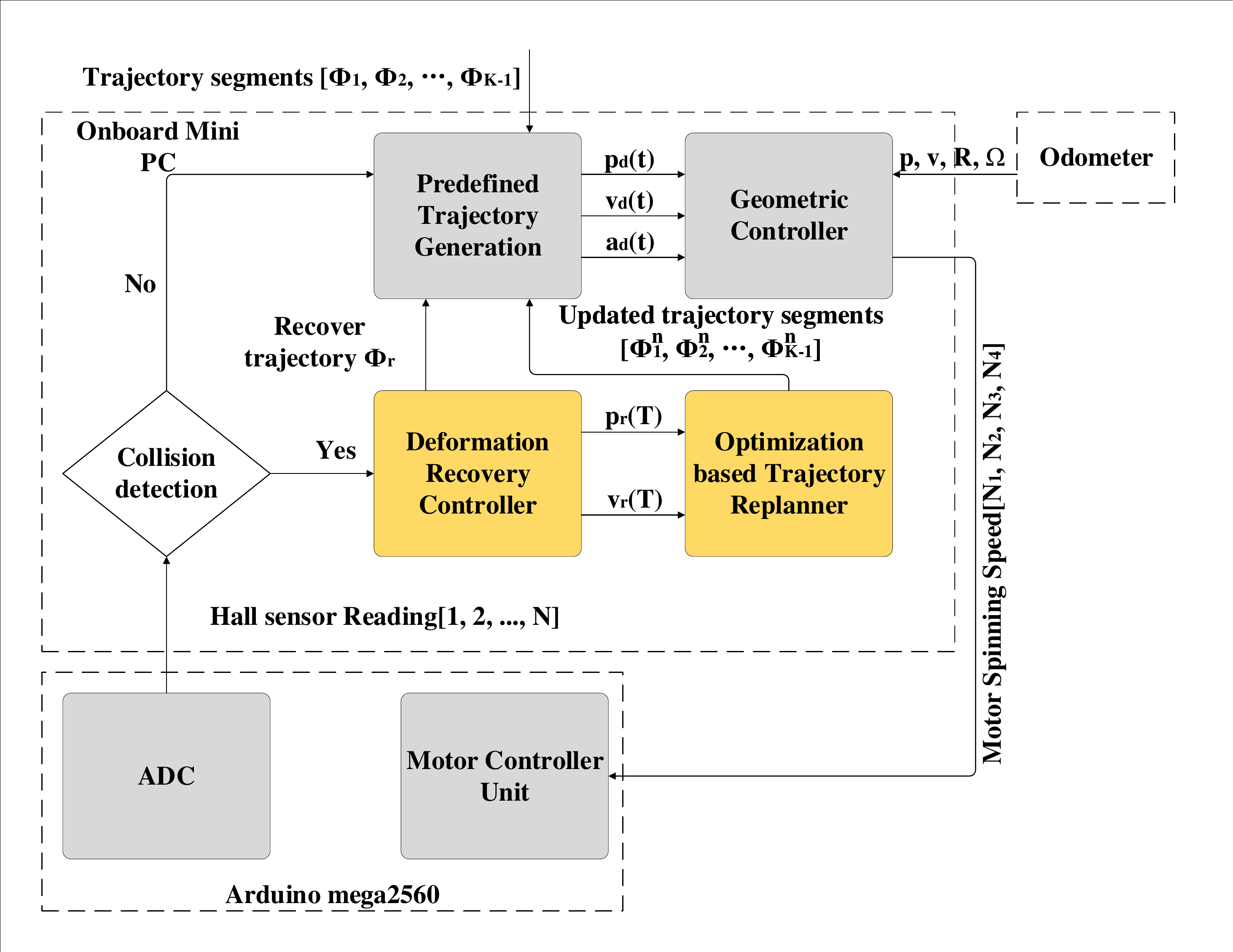}
\vspace{-8pt}
\caption{Overview and software architecture of our DRR strategy for collision-inclusive motion planning and control. The method builds upon two novel components developed in this work; a deformation recovery controller and a post-impact trajectory replanner.}%
\label{fig:Software-architecture}
\vspace{-14pt}
\end{figure}

In contrast to collision avoidance algorithms, we do not impose any obstacle-related constraints in trajectory generation, nor we run a geometric collision check once a trajectory is generated. Instead, we directly generate a trajectory based on given waypoints.\footnote{The list of waypoints can be computed via any path planning method.} If a collision occurs, the robot receives a signal that a collision has occurred from any of the Hall effect sensors embedded between the main chassis and its deflection surfaces and activates a collision recovery controller. The controller (described in Sec.~\ref{sec:deformation}) makes the robot detach from the collision surface and determines a post-collision state for the robot so as to facilitate post-impact trajectory replanning. The replanner (described in Sec.~\ref{sec:replanner}) refines the initial trajectory since collisions change the continuity of derivatives of the trajectory followed before collision. To do so, the replanner uses the post-collision state determined by the recovery controller as the initial state for refined trajectory generation.  
%
The procedure repeats as new collisions may occur in the future, in a reactive and online manner. 
Our proposed \emph{deformation recovery and replanning} (DRR) strategy 
is visualized in Fig.~\ref{fig:Software-architecture}. 

%

%

    
\section{
Deformation Recovery Control Design} \label{sec:deformation}

The purpose of our proposed deformation controller is to make the robot recover from a collision and reach a post-impact state that can facilitate recovery trajectory replanning (which we discuss in the next section). We first present important notation and working assumptions, and then move on to the development of the controller.
    
\subsection{Problem Setting}
\begin{figure}[!ht]
\vspace{6pt}
  \centering\
  \begin{subfigure}{0.235\textwidth}
   \includegraphics[trim={6cm 0.6cm 7.8cm 0.3cm, width=0.74\linewidth}, clip, width=0.95\textwidth]{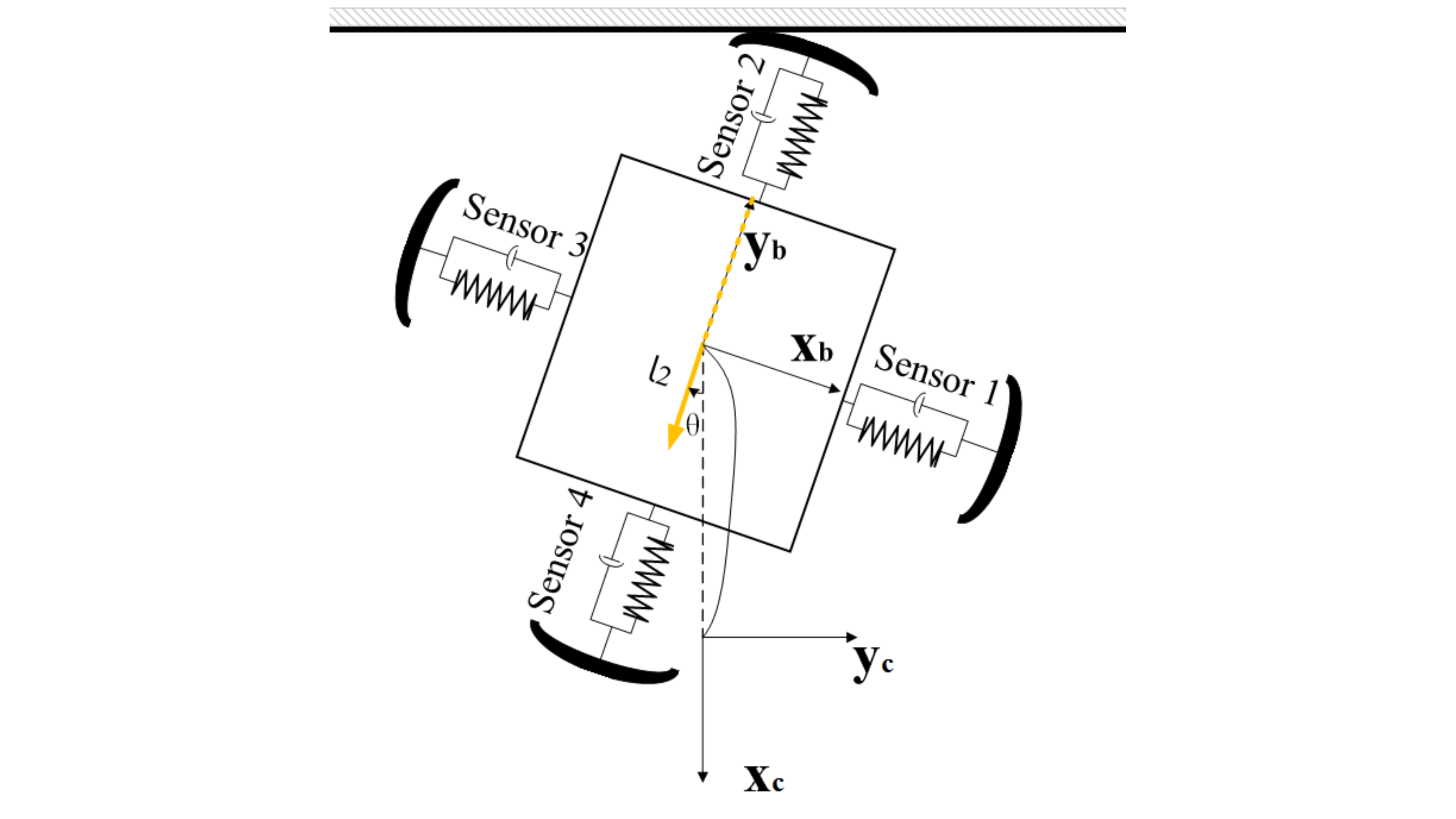}
   \end{subfigure}
    \begin{subfigure}{0.23\textwidth}
        \includegraphics[trim={6cm 0.6cm 7.8cm 0.6cm, width=0.75\linewidth}, clip, width=0.95\textwidth]{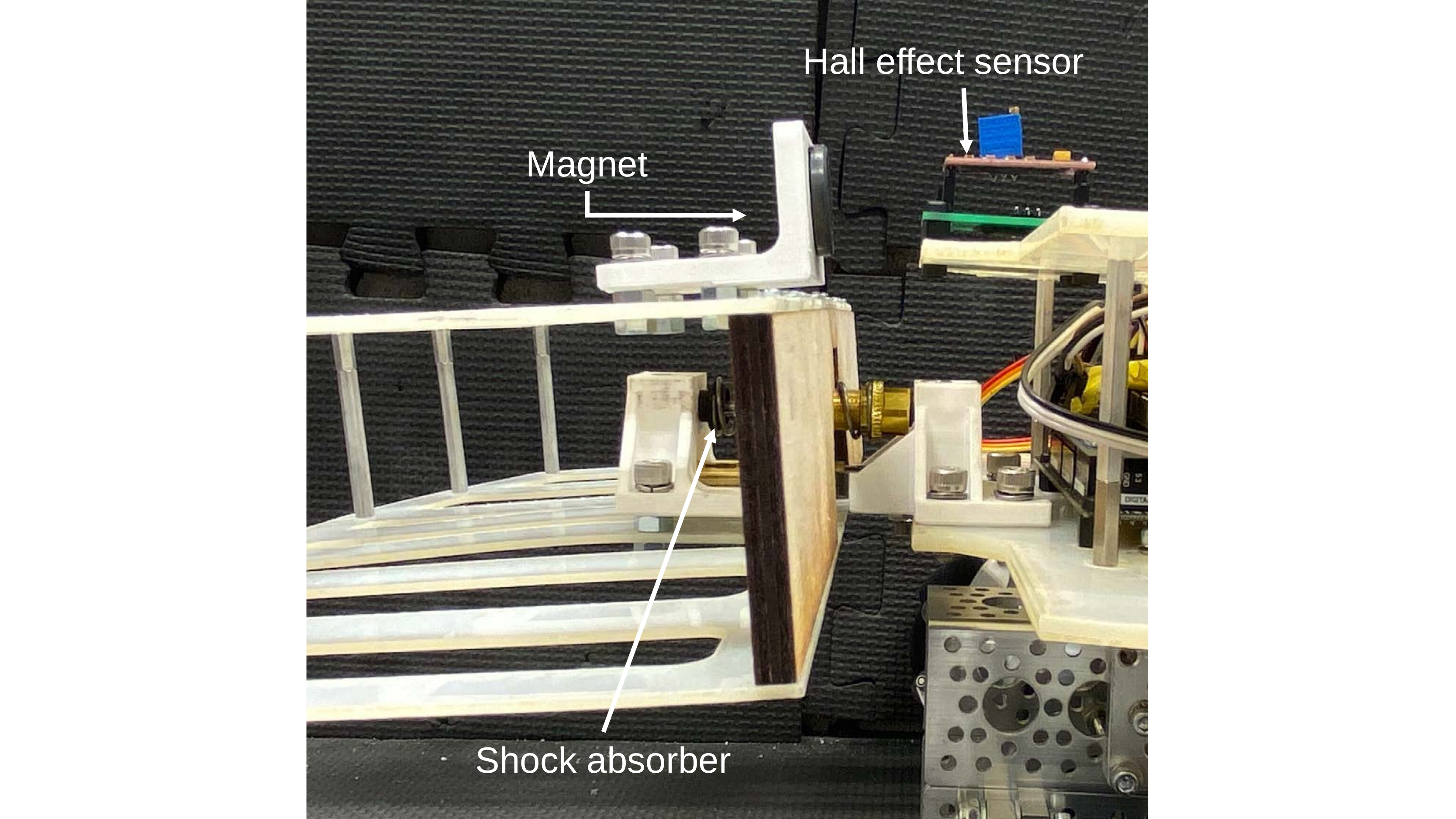}
      \end{subfigure}
      \vspace{-1pt}
\caption{(Left) Model of our wheeled robot equipped with compliant arms. (Right) Close-up view of the assembly of passive visco-elastic prismatic joint and Hall effect sensor.}
\label{fig:Voigt-Collision}
\vspace{-18pt}
\end{figure}

Consider a holonomic mobile robot (Fig.~\ref{fig:introFIG}), modeled as a point mass $m$. 
The robot's main chassis is connected to deflection surfaces via four passive visco-elastic prismatic joints (Fig.~\ref{fig:Voigt-Collision}). Note that the springs inside each passive joint are pre-tensioned.
%
%
The robot's compliant `arms' can both protect the robot from catastrophic impact, and generate a external force driving it away from obstacles. 
External forces along each arm are caused via passive visco-elastic deformations assumed to follow the Voigt model; $k$ and $c$ denote the spring constant and damping coefficient, respectively. 
Hall effect sensors are used to measure the amount of deformation along each arm, and to signal collision detection when a user-tuned arm compression threshold is exceeded.\footnote{The threshold is tuned based on the sensitivity of the Hall effect sensors.}

We consider four key quantities related to spring lengths: neutral $\bm{l}_0$, pre-tensioned $\bm{l}_l$, maximum-load $\bm{l}_e$, and current $\bm{l}$ (also referred to as deformation vector). These quantities play a significant role in the deformation recovery controller; they are also summarized in Table~\ref{tab:notations}, along with other key notation. 
In single-arm collisions, current spring length vector $l$ is aligned with the unit vector along the colliding arm, pointing from the tip of the arm to the center of robot. For clarity of presentation, we consider in the following single-arm collisions. In the case of multi-arm collisions, we compute individual contributions from each colliding arm's spring and then consider their vector sum as the compound deformation vector which is used in lieu of $\bm{l}$.

We use three coordinate systems.  The world and body frames (note $^{w}_{b}\bm{R}$ denotes the rotation matrix from body frame to world frame while $^b\bm{l}$ denotes the deformation vector expressed in the body frame), and a (local) collision frame $\mathcal{F}_c$. This frame is defined at the time instant a collision occurs, $t_c$, remains fixed for the duration of the collision recovery process, $T$, and then it is deleted. Its origin coincides with the origin of the robot when a collision is detected. Basis vector $\{\bm{n}, \bm{t}, \bm{k}\}$ of $\mathcal{F}_{c}$ are defined normal, tangent and upwards with respect to the deformation vector $\bm{l}$. Let $\theta$ be the angle of deformation vector $l$ in $\mathcal{F}_{c}$.

\begin{table}[h!]
\vspace{-4pt}
\caption{List of key notation used in the paper.}
\label{tab:notations}
\vspace{-12pt}
\begin{center}
\begin{tabular}{l l}
\toprule
$\bm{l}_0$&  neutral length of the spring\\
$\bm{l}_s$&  pre-tensioned spring length (arm not compressed)\\
$\bm{l}_e$&  length at maximum spring load following Hooke's law.\\
$\bm{l}$  &  current spring length (deformation vector)\\
\midrule
$k$& spring constant of the arm.\\
$c$& damping coefficient of the arm.\\
\midrule
$~^{w}_{b}\bm{R}$ & Rotation matrix from body frame to world frame\vspace{2pt}\\
$~^{w}_{c}\bm{R}$ & Rotation matrix from collision frame $\mathcal{F}_c$ to world frame\\
\bottomrule
\end{tabular}
\end{center}
\vspace{-14pt}
\end{table}

We consider planar collision cases with polygon-shaped obstacles and assume that during deformation and until the collided arm recovers its initial length: 1) the tip of the arm remains in contact with the obstacle's collision surface 
but does not rotate about the $z$ axis, and 2) the wheels of the robot contact the ground. %
%
%
%
The (frame-agnostic) robot collision dynamics is then given by \[m\ddot{\bm{l}} + c\dot{\bm{l}} +k(\bm{l} - \bm{l_{0}}) = m{\bm{a}}_{in},\]
where ${\bm{a}}_{in}$ is the robot's body acceleration input. 

\subsection{Deformation Controller}\label{subsec:deformation}
The main idea underlying the proposed collision deformation recovery controller is to steer the post-impact state of the robot to a desired one within a time period of $[t_{c}, t_{c} + T]$. The time horizon $T$ is an important hyper-parameter tuned by the user. Typically, longer $T$ means the robot will recover from collision with longer time and smoother motion pattern; in this paper, we select $T = 0.5$\;s.

We solve this problem by generating the state-space model of this problem first. Then we linearize the nonlinear state-space model using feedback linearization. We formulate a fixed-horizon ($T$) optimal control problem to solve for the control input of the linearized system. Finally, we discretize the linearized system fixed-horizon control problem and formulate it as a constrained quadratic programming problem.


The deformation controller operates with respect to the local, collision frame $\mathcal{F}_{c}$. Let the 
state variable be $~^{c}\bm{x} = [x\ y\ \theta \  v_{x}\ v_{y}]^{\top}$. 
The control input is $\bm{u} = [u_{x}\ u_{y}\  u_{\theta}]^{\top}$, where
$u_{x} = (^c\bm{a}_{in} - \frac{k}{m}(^c\bm{l}_{s} - \hspace{-3pt}~^c\bm{l}_{0}))\cdot\hspace{-3pt}~^c\bm{n}$, 
$u_{y} = (^c\bm{a}_{in} - \frac{k}{m}(^c\bm{l}_{s} - \hspace{-3pt}~^c\bm{l}_{0}))\cdot\hspace{-3pt}~^c\bm{t}$, and 
$u_{\theta} =\hspace{1pt} ^c\hspace{-0pt}\bm{\omega}\cdot\hspace{-3pt}~^c\bm{k}$ with $^c\bm{\omega}$ being the angular velocity of the robot in the collision frame. Note that position control terms include compensation for the force caused by the spring being pre-tensioned when the robot's arm is at its rest length.
%
%
Then, the state space model of the robot recovering from collision can be expressed as
\begin{equation}\label{state space model}
    \left\{
    \begin{array}{lr} 
        \dot{x} = v_{x}& \\
        \dot{y} = v_{y}& \\
        \dot{\theta} = u_{\theta}& \\
        \dot{v}_{x} = -\frac{k}{m}{x} - \frac{c}{m}{v_{x}} + u_{x}& \\
        \dot{v}_{y} = -\frac{k(\mu \sgn(v_{y}) + \tan \theta)x + f_{0}}{m} -\frac{c(\mu \sgn(v_{y}) + \tan \theta)v_{x}}{m} + u_{y}
    \end{array} \right.
\end{equation}
where $f_{0} = \mu k\sgn(v_{y}) (^c\bm{l}_{s} - \hspace{-3pt}~^c\bm{l}_{0})\cdot\hspace{-3pt}~^c\bm{n}$. 

Since the robot is holonomic, we can decouple orientation from position control. In our approach we seek to make the robot keep the same orientation it has at the instant it collides throughout the collision recovery process. We follow this approach because it can simplify the overall deformation recovery control problem without sacrificing optimality.

The orientation and angular velocity errors during recovery time $t\in[t_c,t_c+T]$ are $\bm{e}_{R}(t) = \frac{1}{2}(\bm{R}^{\top}_{d}\bm{R}-\bm{R}^{\top}\bm{R}_{d})^{\vee}$ and $\bm{e}_{\dot{R}}(t) = \bm{\omega} - \bm{R}^{\top}\bm{R}_{d}\bm{\omega}_{d}$, respectively.\footnote{The vee map $\vee$ is the inverse of a skew-symmetric mapping.} 
Index $d$ denotes desired quantities; these are $\bm{R}_{d} = \bm{R}(t_{c})$ and $\bm{\omega}_{d} = [0\ 0\ 0]^{\top}$. (All terms are with respect to collision frame $\mathcal{F}_c$.)
Then,  
\begin{equation}\label{eq:rotational}
u_{\theta} = -K_{r}e_{R, z}(t)-K_{\omega}e_{\dot{R}, z}(t)\enspace.
\end{equation}
Note that since this is a planar collision problem, the collision recovery orientation controller considers only the $z-$components of orientation and angular velocity errors. 


We now turn our attention to collision recovery position control. 
The translation-only motion in \eqref{state space model} is affine. Therefore, we can apply feedback linearization. 
The linearized system matrix $\bm{F}$ is
\begin{align*}
 \bm{F} = \begin{bmatrix}
0 & 0 & 1 & 0 \\
0 & 0 & 0 & 1\\
-\frac{k}{m} & 0 & -\frac{c}{m} & 0\\
0  & 0 & 0 & 0\\
\end{bmatrix}
\end{align*}
with state vector $\bm{s} = [x\ y\ v_{x}\ v_{y}]^{\top}$. The control input matrix is $\bm{G} = \bm{I_{2\times2}}$ with control input vector $\bm{\nu} = [\nu_{x}\ \nu_{y}]$ given by 
\begin{equation} \label{eq:feedback linearization}
\left\{
    \begin{array}{lr} 
        \nu_{x} = u_{x}& \\
        \nu_{y} = u_{y} - \frac{k(\mu \sgn(v_{y}) + \tan \theta)x + f_{0}}{m} - \frac{c(\mu \sgn(v_{y}) + \tan \theta)v_{x}}{m}
\end{array} 
    \right.\hspace{-16pt}
\end{equation}

We formulate an optimal control problem with fixed time horizon $T$ based on the linearized system $\dot{\bm{s}} = \bm{F}\bm{s} + \bm{G}\bm{\nu}$. Using the change of variable $\tau=t-t_c$,\footnote{We employ this change of variable for clarity. Problem (4) resets every time a new collision occurs; this gives rise to an LTI system, hence the change of variable can apply.} we seek to solve
    \begin{subequations}\label{optimal control}
    \begin{alignat}{2}
    &\!\min_{\bm{s}}        &\qquad& \int\limits_{0}^{  T}{(\bm{s}(\tau)^{\top}\bm{\Gamma}\bm{s}(\tau)+\bm{\nu}^{\top}(\tau)\bm{H}\bm{\nu}(\tau))}d\tau\label{eq:optctrl}\\
    &\text{subject to} &      & \dot{\bm{s}} = \bm{F}\bm{s} + \bm{G} \bm{v},\label{eq:constraint1}\\
    &                  &      & -\lVert \bm{l}_{e} - \bm{l}_{s} \rVert \cos{\theta}\leq x \leq 0.\label{eq:constraint2}\\
    &                  &      & \bm{s}(0) = [x_{0}\ 0\ v_{0,x}\
    v_{0,y}].\label{eq:constraint3}\\
    &                  &      & \bm{s}(T) = [0\ y_{T}\ v_{T,x}\
    v_{T,y}].\label{eq:constraint4}
    \end{alignat}
    \end{subequations}

Matrices $\Gamma = \gamma\begin{bmatrix}
    \bm{I_{2\times2}} & \bm{0} \\
    \bm{0} & \bm{0_{2\times2}} \\
\end{bmatrix}$ and $H = h\bm{I_{2\times2}}$ 
penalize the displacement during the recovery process and the control input, respectively. There is a trade-off between the displacement and the control input of the robot. Tuning parameters $\gamma$ and $h$ balance this trade-off to select the controller with minimal control energy and displacement.

Constraint~\eqref{eq:constraint2} dictates that the robot should be in contact with the collision surface until the colliding arm's spring has recovered its original, pre-tensioned length $l_s$ (i.e. the arm is no longer compressed) without compressing beyond its linear region $l_e$. 
Constraints~\eqref{eq:constraint3} and~\eqref{eq:constraint4} enforce initial and terminal position and velocity conditions, respectively. 
In detail, $x_0$ is determined by the colliding arm's Hall effector sensor reading. Since the vector form of the sensor's reading (that is, $~^{b}\bm{l}-~^{b}\bm{l}_s$) is expressed in the body frame, we need to transform it to the collision frame $\mathcal{F}_{c}$ as per
\begin{equation}\label{eq: x0}
    x_{0} = -[1\;0] 
    ~^{w}_{c}\bm{R}^\top 
    ~^{w}_{b}\bm{R}
    ~(~\hspace{-3pt}^{b}\bm{l} - \hspace{-3pt} ~^{b}\bm{l}_{s})\enspace.
\end{equation}

The velocity components at the collision instant $v_{0,x}$ and $v_{0,y}$ are expressed in frame $\mathcal{F}_{c}$ and are estimated at run-time.\footnote{In the experiments conducted in this work, velocity measurements are provided via a motion capture camera system, but the method can apply as long as velocity estimates are available, e.g., via optical flow.} 
Post-impact, the arm needs to be uncompressed (hence $x_T$ is set to $0$), but $y_{T}$ is left as an unconstrained free variable. 
Post-impact terminal velocity components $v_{T,x}$ and $v_{T,y}$ are also expressed in $\mathcal{F}_{c}$ and can be set freely. In Alg.~\ref{recovery controller} lines 1--8, we discuss how to generate $v_{T,x}$ and $v_{T,y}$ based on the preplanned trajectory.
We discretize the linearized system \eqref{eq:constraint1} with sampling frequency $f=10$\;Hz using the Euler method, and solve the corresponding quadratic program with CVXOPT. The process is summarized in Alg.~\ref{recovery controller}. 

Computed control inputs~\eqref{optimal control} and~\eqref{eq:rotational} make the robot detach from the collision surface and help bring it to a temporary post-collision state which can be used as the initial condition for post-impact trajectory generation. We discuss this next.

\begin{algorithm}[h!]
	\caption{Collision recovery control algorithm.}
	\label{recovery controller}
	\LinesNumbered
	\SetKwInOut{Input}{input}
	\SetKwInOut{Output}{output}
	\SetKwInOut{Parameter}{parameter}
	\Input{Displacement in body frame $~^{b}\bm{l} - ~^{b}\bm{l}_{s}$ via Hall effect sensors readings; 
	collision time instant $\tau_{c} \in [0, \Delta t_{i_{c}})$; position in world frame at collision instant, $~^{w}\bm{p}_{\tau_{c}}$; velocity in world frame at collision instant, $~^{w}\bm{v}_{\tau_{c}}$; rotation matrix 
	$~^{w}_{b}\bm{R}$; 
	rotation matrix 
	$~^{w}_{c}\bm{R}$; next waypoint point in world frame, $~^{w}\bm{p}_{next}$.}
	\Output{Control input $\bm{u}$}
	\Parameter{Maximum velocity of the robot $v_{max}$}
	    $~^{w}\bm{v}_{T} \leftarrow \frac{~^{w}\bm{p}_{next} - ~^{w}\bm{p}_{\tau_{c}}} {\Delta t_{i_{c}} - \tau_{c}}$\\
	     $~^{c}\bm{v}_{T} \leftarrow ~^{w}_{c}\bm{R}^{\top}~^{w}\bm{v}_{T}$\\
	    \If{$~^{c}\bm{v}_{T, x} < 0$}
	    {$~^{c}\bm{v}_{T, x} \leftarrow 0$\\
	    }
	    \If{$\lVert ~^{c}\bm{v}_{T} \rVert \geq v_{max}$}
	    {$~^{c}\bm{v}_{T, y} \leftarrow v_{max}normalize(~^{c}\bm{v}_{T})$
	    }
	    Calculate $x_{0}$ based on \eqref{eq: x0} with $\bm{l}_{b}$\\
	    $v_{T, x} \leftarrow ~^{c}\bm{v}_{T, x}$, $v_{T, y} \leftarrow ~^{c}\bm{v}_{T, y}$\\
	    Calculate $u_{x}$, $u_{y}$ based on \eqref{optimal control} and \eqref{eq:feedback linearization} with given $v_{T, x}$, $v_{T, y}$ and $x_{0}$\\
	    Calculate $u_{\theta}$ based on \eqref{eq:rotational}\\
	\Return $\bm{u} \leftarrow [u_{x}\ u_{y}\  u_{\theta}]^{\top}$
\end{algorithm}

\section{Post-impact Trajectory Replanning}\label{sec:replanner}

%
We formulate the post-impact trajectory generation problem as a quadratic program with equality constraints. Letting $J'=\sum\limits_{i = i_{c}}^{I}~^{w}\bm{q}_{i, \beta}^{\top}~^{w}Q^{j}_{\beta}(\Delta t_{i})~^{w}\bm{q}_{i, \beta}$, we seek to solve 
\vspace{-6pt}
\begin{subequations} \label{Trajectory}
\begin{alignat}{2}
&\!\min_{\bm{q}}   &\qquad& 
   J(\bm{q}) = \sum\limits_{i = i_{c}}^{I}\int\limits_{0}^{\Delta t_{i}}{\left\lVert ~^{w}P_{i, \beta}^{j}(t)\right\rVert}dt = J' 
\label{eq:minimal_effort}\\
&\text{subject to} &      & ~^{w}\bm{A}^{(0)}_{0, i_{c}, \beta}\bm{q}_{i_{c}, \beta} = ~^{w}\bm{p}_{r, \beta}, 
\label{eq:me_constraint1}\\
&                  &      &  ~^{w}\bm{A}^{(1)}_{0, i_{c}, \beta}\bm{q}_{i_{c}, \beta} = ~^{w}\bm{v}_{r, \beta}, 
\label{eq:me_constraint2}\\
&                  &      &  ~^{w}\bm{A}^{(\alpha)}_{\Delta t_{I}, I, \beta}\bm{q}_{I, \beta} = ~^{w}\bm{d}_{\Delta t_{I}, I, \beta}^{(\alpha)}, 
\label{eq:me_constraint3}\\
&                  &      &  ~^{w}\bm{A}^{0}_{\Delta t_{i}, i, \beta}\bm{q}_{i, \beta} = ~^{w}\bm{p}_{i+1, \beta}, 
\label{eq:me_constraint4}\\
&                  &      & ~^{w}\bm{A}_{\Delta t_{i},i+1}^{(\alpha)}\bm{q}_{i, \beta} = \bm{A}_{0,i+1, \beta}^{(\alpha)}\bm{q}_{i+1, \beta}.\label{eq:me_constraint5}
\end{alignat}
\end{subequations}

Superscript $j$ denotes the derivative order; for example, $j = \{1, 2, 3, 4\}$ correspond to min-velocity, min-acceleration, min-jerk and min-snap trajectories, respectively. $\beta \in \{x, y\}$ indicates the $x$ and $y$ component of the trajectory. 
$\Delta t_{i}$ is the time duration for $i^{th}$ polynomial segment. 
Parameter $\bm{q}_{i, \beta}$ is the vector of coefficients of $i^{th}$ polynomial. 
$~^{w}\bm{A}_{0, i, \beta}^{(\alpha)}$ maps the coefficients to $\alpha^{th}$ order derivative of the start point in segment $i$, while $~^{w}\bm{A}_{\Delta t_{i}, i, \beta}^{(\alpha)}$ maps the coefficients to $\alpha^{th}$ order derivative of the end point in segment $i$, for $\alpha \in \{0, 1, \ ...\ j-1 \}$. 
Constraints~\eqref{eq:me_constraint1} and~\eqref{eq:me_constraint2} impose the initial values for the $0^{th}$ and the $1^{st}$ order derivatives to match the position and velocity values attained via the collision recovery controller, respectively. 
Constraint~\eqref{eq:me_constraint3} imposes that the $\alpha^{th}$ order derivatives of the end position are fixed. 
Constraint~\eqref{eq:me_constraint4} imposes that the trajectory will pass through desired waypoints after $i_{c}$. 
Constraint~\eqref{eq:me_constraint5} is imposed to ensure $\alpha^{th}$ continuity among polynomial segments. 
%

We solve this QP problem given initial (post-collision) and end states, and intermediate waypoints. 
Then we perform time scaling as in~\cite{liu2017planning} to reduce the maximum values for planned velocities and accelerations, as well as higher-order derivatives as appropriate, and thus improve dynamic feasibility of the refined post-impact trajectory.
    

\subsection{Waypoint Adjustment}
In some cases, we need to adjust the waypoints given in a preplanned trajectory with the information we get from the collision and then solve~\eqref{Trajectory} with the adjusted waypoints. Such cases occur when 
there is no direct line of sight between the collision state and the waypoint at the end of the immediately next trajectory segment following collision recovery. 
By enabling such waypoint adjustment, the algorithm promotes exploration and in certain cases prevents the robot from being trapped in a local minima in which repeated collisions at the same (or very close-by) place could otherwise occur. 

With reference to Alg.~\ref{pseudo replanner}, we express in the local collision frame $\mathcal{F}_c$ the next waypoint $~^{w}\mathbf{waypoint\_list}[i_{c} + 1]$ (lines 1--3). In lines 4--8, we adjust the waypoint at the end of $i_{c}$ segment if it lies on the same direction of $~^{c}\bm{p}_{r}$ by moving it so to enable direct line of sight to collision state. In lines 9--17, if the waypoint at the end of $i_{c}$ segment lies on the opposite direction of $~^{c}\bm{p}_{r}$, 
we insert a new intermediate waypoint in the waypoint list. The waypoint is generated by displacing $^{c}\bm{p}_{r}$ for a fixed (user-defined) `exploration distance' $\epsilon_{explore}$ expressed in $\mathcal{F}_{c}$. In lines 18--23 we insert a new waypoint in the list as in lines 13--16 when there is direct line of sight with the waypoint at the end of $i_{c}$ segment but the waypoint lies on the opposite direction of $^{c}\bm{p}_{r}$.

    
\vspace{-9pt}
\begin{algorithm}[h!]
	\caption{Waypoint adjustment algorithm}
	\label{pseudo replanner}
	\LinesNumbered
	\SetKwInOut{Input}{input}
	\SetKwInOut{Output}{output}
	\SetKwInOut{Parameter}{parameter}
	\Input{Displacement $~^{b}\bm{l} - ~^{b}\bm{l}_{s}$; Position after the collision recovery in world frame, $~^{w}\bm{p}_{r}$; waypoint list of preplanned trajectory;  $~^{w}\mathbf{waypoint\_list}$;  $~^{w}_{c}\bm{R}$; trajectory segment $i_{c}$ where the collision happens.}
	\Output{waypoint list after adjustment $~^{w}\mathbf{waypoint\_list}$}
	\Parameter{Robot radius $\rho$}
	$~^{w}\bm{p}_{next} \leftarrow ~^{w}\mathbf{waypoint\_list}[i_{c} + 1]$ \\
	Transfer $~^{w}\bm{p}_{next}$ into $\mathcal{F}_{c}$ frame to get $~^{c}\bm{p}_{next}$\\
	Transfer $~^{w}\bm{p}_{r}$ into $\mathcal{F}_{c}$ frame to get $~^{c}\bm{p}_{r}$\\
	    \If{$-\rho \leq ~^{c}\bm{p}_{next, x} < ~^{c}\bm{p}_{r, x}$}
	    {$~^{c}\bm{p}_{next, x} \leftarrow ~^{c}\bm{p}_{r, x}$\\
	    Transfer $~^{c}\bm{p}_{next}$ into world frame to get $~^{w}\bm{p}_{next}$\\
	    $~^{w}\mathbf{waypoint\_list}[i_{c} + 1] \leftarrow ~^{w}\bm{p}_{next}$
	    }
	    \If{$-2\rho \leq ~^{c}\bm{p}_{next, x} < -\rho$}
	    {$~^{c}\bm{p}_{next, x} \leftarrow -2\rho$\\
	    Transfer $~^{c}\bm{p}_{next}$ into world frame to get $~^{w}\bm{p}_{next}$\\
	    $~^{w}\mathbf{waypoint\_list}[i_{c} + 1] \leftarrow ~^{w}\bm{p}_{next}$\\
	    $~^{c}\bm{p}_{add, x} \leftarrow ~^{c}\bm{p}_{r, x}$\\
	    $~^{c}\bm{p}_{add, y} \leftarrow ~^{c}\bm{p}_{r, y} + \epsilon_{explore}$\\
	    Transfer $~^{c}\bm{p}_{add}$ into world frame to get $~^{w}\bm{p}_{add}$\\
	    Insert a waypoint $~^{w}\bm{p}_{add}$ between $i_{c}$ and $i_{c} + 1$
	    }
	    \If{$ ~^{c}\bm{p}_{next, x} < -2\rho$}
	    {$~^{c}\bm{p}_{add, x} \leftarrow ~^{c}\bm{p}_{r, x}$\\
	    $~^{c}\bm{p}_{add, y} \leftarrow ~^{c}\bm{p}_{r, y} + \epsilon_{explore}$\\
	    Transfer $~^{c}\bm{p}_{add}$ into world frame to get $~^{w}\bm{p}_{add}$\\
	    Insert a waypoint $~^{w}\bm{p}_{add}$ between $i_{c}$ and $i_{c} + 1$
	    }
	\Return $~^{c}\mathbf{waypoint\_list}$
\end{algorithm}
\vspace{-9pt}

If a new waypoint is inserted in the list, we map the path 
generated by $^{w}\bm{p}_{r}$ and waypoints in the list after $i_{c} + 1$ into time domain using a trapezoidal velocity profile. If no new waypoint is inserted, we set the time duration of $i_{c}$ segment in \eqref{Trajectory} as $\Delta t_{i_{c}} = t_{i_{c}+1} - t_{c}$, where $t_{i_{c}+1}$ is the time reaching the next waypoint $\bm{p}_{i+1}$ per the preplanned trajectory.

\section{Experimental Results}
\subsection{Robot and Environment Setup}
We test our proposed algorithm with an omni-directional impact-resilient wheeled robot we built in-house (Fig.~\ref{fig:introFIG}). The main chassis is connected to a deflection `ring' via four arms that feature a passive visco-elastic prismatic joint each.  
Each passive arm has embedded Hall effect sensors to measure the length of the arm and detect collisions along each of their direction when the deformation exceeds a certain threshold. 

Odometry feedback is provided by a 12-camera VICON motion capture system. The robot operates in an $2.0m \times 2.0m$ area with a rectangular pillar serving as a static polygon-shaped obstacles. An onboard Intel NUC mini PC ($2.3$\;GHz i7 CPU; $16$\;GB RAM) processes odometry data and sends control commands to the robot at a frequency of $10$\;Hz.

The robot may flip when colliding with a velocity over an upper bound. To identify a theoretical collision velocity bound to avoid flipping, we use an energy conservation argument. Assume the kinetic energy before collision transfers into elastic potential energy of the passive arm, and the gravitational potential energy of the robot with small flipping angle counters the negative work input from the controller: 
\vspace{-10pt}

{\small
    {\begin{equation*}
        E_{k, t^{-}}(v_{max}) \hspace{-2pt}=\hspace{-2pt} E_{ep}(l_{e}) - E_{ep}(l_{s}) + E_{gp}(\sigma_{max}) + ma_{in, max}(l_{s} - l_{e}).
    \end{equation*}
    }%
}
Then,
\vspace{-6pt}
{\small{
\begin{equation*}
\begin{aligned}
    v_{max} = & \{\frac{k[(l_{e} - l_{0})^{2} - (l_{s}-l_{0})^{2}]}{2m} + g(\rho - l_{s} + l_{e})\sin{\sigma_{max}}\\
    + & a_{in, max}(l_{s}-l_{e})\}^{\frac{1}{2}} \enspace. 
\end{aligned} 
\end{equation*}
}%
}
The robot's radius is $\rho=0.3$\;m. The difference between the initial and neutral position of each passive arm is $l_{s} = 30.0$\;mm; the maximum load length is $l_{e} = 15.0$\;mm; and the neutral length is $l_{0} = 41.5$\;mm. The spring coefficient $k = 2.31$\;N/mm. We select the largest flip angle $\sigma_{max} = \ang{3}$. The maximum acceleration input from the robot is $a_{in, max} = 5.0$\;m/s$^{2}$. The mass of the robot is $6.0$\;kg. Then, we compute an upper theoretical velocity bound of $v_{max} \approx 0.7$\;m/s. 
    
    
The structure and number of passive arms affect collision detection. The most accurate detection occurs when the impact is located along the direction of arm. Collision detection accuracy can thus increase by adding more passive arms. Here, we use four arms to demonstrate how our proposed algorithm works, but future iterations will consider more arms, especially as we operate in more cluttered environments.

\begin{table*}[!h]
\vspace{6pt}
    \caption{Deformation controller testing results.}
    \vspace{-10pt}
    \label{table:improvement}
    \begin{center}
    \begin{tabular}{l l l l l l}
    \toprule
    $\bm{v}_{in,set}$ & $\bm{v}_{out,set}$ & $\bar{\bm{v}}_{in}$ & $STD (\bm{v}_{in})$ & $\bar{\bm{v}}_{out}$ & STD ($\bm{v}_{out}$)\\
    \toprule
    $(0.0,0.5)$ 
    & $(0.0,-0.5)$
    & $(0.017,0.532)$
    & $(0.013, 0.028)$
    & $(0.005,-0.372)$
    & $(0.019,0.189)$\\
    \midrule
     $(0.0,0.7)$ 
    & $(0.0,-0.5)$
    & $(0.040,0.709)$
    & $(0.015, 0.055)$
    & $(0.005,-0.387)$
    & $(0.021, 0.223)$\\
    \midrule
    $(0.0,0.5)$ 
    & $(0.0,-0.7)$
    & $(0.010,0.500)$
    & $(0.012, 0.027)$
    & $(0.005,-0.340)$
    & $(0.030, 0.155)$\\
    \midrule
    $(0.0,0.7)$ 
    & $(0.0,-0.7)$
    & $(0.036,0.709)$
    & $(0.016, 0.055)$
    & $(-0.016,-0.371)$
    & $(0.029, 0.223)$\\
    \midrule
    \midrule
    $(0.0,0.5)$ 
    & $(0.345,-0.345)$
    & $(0.022,0.509)$
    & $(0.012, 0.027)$
    & $(0.017,-0.329)$
    & $(0.037, 0.213)$\\
    \midrule
    $(0.0,0.7)$ 
    & $(0.345,-0.345)$
    & $(0.034,0.691)$
    & $(0.020, 0.053)$
    & $(0.010,-0.327)$
    & $(0.029, 0.201)$\\
    \midrule
    $(0.0,0.5)$ 
    & $(0.495,-0.495)$
    & $(0.023,0.511)$
    & $(0.022, 0.040)$
    & $(0.027,-0.259)$
    & $(0.066, 0.192)$\\
    \midrule
     $(0.0,0.7)$ 
    & $(0.495,-0.495)$
    & $(0.027,0.733)$
    & $(0.011, 0.056)$
    & $(0.046,-0.293)$
    & $(0.082, 0.305)$\\
    \midrule
    \midrule
    $(0.0,0.5)$ 
    & $(0.5,0.0)$
    & $(0.032,0.496)$
    & $(0.020, 0.047)$
    & $(0.016,-0.272)$
    & $(0.035, 0.188)$\\
    \midrule
    $(0.0,0.7)$ 
    & $(0.5,0.0)$
    & $(0.036,0.697)$
    & $(0.011, 0.062)$
    & $(0.033,-0.334)$
    & $(0.073, 0.248)$\\
    \midrule
    $(0.0,0.5)$ 
    & $(0.7,0.0)$
    & $(0.017,0.517)$
    & $(0.019, 0.015)$
    & $(0.048,-0.300)$
    & $(0.057, 0.217)$\\
    \midrule
     $(0.0,0.7)$ 
    & $(0.7,0.0)$
    & $(0.026,0.722)$
    & $(0.009, 0.024)$
    & $(0.014,-0.261)$
    & $(0.045, 0.219)$\\
    \midrule
    \midrule
    $(0.345,0.345)$ 
    & $(0.0,-0.5)$
    & $(0.313,0.380)$
    & $(0.023, 0.025)$
    & $(-0.007,-0.113)$
    & $(0.043, 0.080)$\\
    \midrule
    $(0.495,0.495)$ 
    & $(0.0,-0.5)$
    & $(0.403,0.485)$
    & $(0.048, 0.038)$
    & $(0.002,-0.126)$
    & $(0.027, 0.086)$\\
    \midrule
    $(0.345,0.345)$ 
    & $(0.0,-0.7)$
    & $(0.311,0.372)$
    & $(0.020, 0.041)$
    & $(-0.019,-0.114)$
    & $(0.046, 0.086)$\\
    \midrule
     $(0.495,0.495)$ 
    & $(0.0,-0.7)$
    & $(0.417,0.497)$
    & $(0.047, 0.034)$
    & $(0.010,-0.107)$
    & $(0.045, 0.086)$\\
    \midrule
    \midrule
    $(0.345,0.345)$ 
    & $(0.345,-0.345)$
    & $(0.281,0.351)$
    & $(0.022, 0.014)$
    & $(-0.006,-0.061)$
    & $(0.040, 0.055)$\\
    \midrule
    $(0.495,0.495)$ 
    & $(0.345,-0.345)$
    & $(0.346,0.472)$
    & $(0.052, 0.033)$
    & $(0.002,-0.166)$
    & $(0.042, 0.077)$\\
    \midrule
    $(0.345,0.345)$ 
    & $(0.495,-0.495)$
    & $(0.293,0.367)$
    & $(0.036, 0.029)$
    & $(-0.009,-0.099)$
    & $(0.047, 0.060)$\\
    \midrule
     $(0.495,0.495)$ 
    & $(0.495,-0.495)$
    & $(0.353,0.462)$
    & $(0.034, 0.036)$
    & $(0.006,-0.125)$
    & $(0.045, 0.086)$\\
    \midrule
    \midrule
    $(0.345,0.345)$ 
    & $(0.5,0.0)$
    & $(0.261,0.345)$
    & $(0.023, 0.038)$
    & $(0.023,-0.095)$
    & $(0.056, 0.099)$\\
    \midrule
    $(0.495,0.495)$ 
    & $(0.5,0.0)$
    & $(0.394,0.492)$
    & $(0.079, 0.129)$
    & $(0.031,-0.140)$
    & $(0.027, 0.100)$\\
    \midrule
    $(0.345,0.345)$ 
    & $(0.7,0.0)$
    & $(0.309,0.390)$
    & $(0.127, 0.166)$
    & $(0.013,-0.035)$
    & $(0.012, 0.030)$\\
    \midrule
     $(0.495,0.495)$ 
    & $(0.7,0.0)$
    & $(0.351,0.471)$
    & $(0.025, 0.040)$
    & $(0.002,-0.17)$
    & $(0.038, 0.108)$\\
\bottomrule
    \end{tabular}
    \end{center}
    \vspace{-9pt}
\end{table*}

\subsection{Experimental Testing of the Deformation Controller}\label{subsect:deformation controller}
To examine the deformation controller's effect in the overall trajectory generation method, we command the robot to collide with an obstacle and then apply the proposed deformation recovery controller. 
%
We perform $10$ trials of each combination of input and output velocities (see Table~\ref{table:improvement}; variables with bars denote averages of observed values). 
We note that there are very few cases that collisions were not detected; only $9$ out of $249$. 

Results suggest that the deformation controller generates a negative velocity to make the robot detach from the obstacle after collision. Actual output velocity $\bar{\bm{v}}_{out}$ is determined by the actual input velocity $\bar{\bm{v}}_{in}$ and the set output value $\bm{v}_{out,set}$ though the latter may not be reached in practice. That is because feedback linearization is not robust to system parameter uncertainties that occur in practice.
We observe that the velocity along $\bm{n}$ is closer to the set velocity than the velocity along $\bm{t}$. This is because most of the uncertainties in system parameters enter as unmodeled friction dynamics along the $\bm{t}$ axis. In future iterations we will upgrade the controller and the sensor setup to make the robot track the predefined velocity along $\bm{t}$ more accurately as well. However, it is sufficient for the recovery velocity to make the robot detach from the obstacle (which it does).
%
Further, the sensor is more accurate when the input velocity is along $\bm{n}$; the average value of deformation detected is $29\%$ larger.  
\subsection{Experimental Testing of the Overall DRR Strategy}

We test our proposed DRR strategy with a trajectory generated based on solving an unconstrained QP problem and following time allocation and time scaling as in~\cite{liu2017planning} without collision checking. We compare our algorithm's performance against the trajectory generation strategy in~\cite{richter2016polynomial} with time allocation and scaling as in~\cite{liu2017planning}. We compare the strategies under two cases: 1) when the previous path does not intersect with the collision surface; and 2) when the previous path intersects with the collision surface.

\begin{figure}[!t]
\vspace{0pt}
\centering
\includegraphics[trim={0.15cm 0.25cm 1cm 0.85cm}, clip, width=0.325\textwidth]{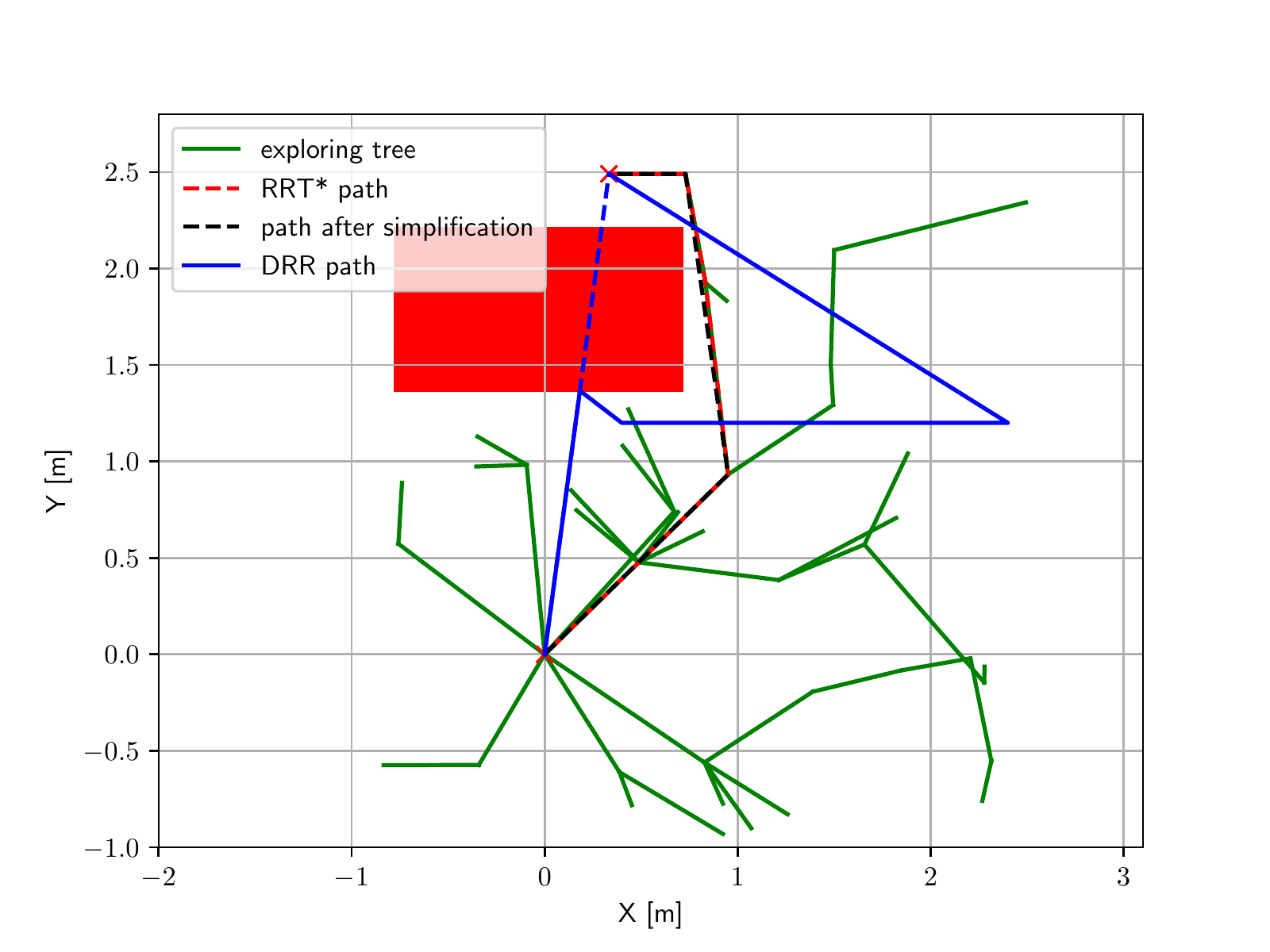}
\vspace{-3pt}
\caption{An example of replanned path in case 2 using proposed DRR strategy and collision avoidance RRT* algorithm in~\cite{richter2016polynomial}.
      }
      \label{fig:path_in_case_2}
	\vspace{-15pt}
\end{figure}

\begin{figure}[!h]
\vspace{-2pt}
      \centering
      \begin{subfigure}{0.235\textwidth}
        \includegraphics[trim={6cm 0.2cm 6cm 5.5cm}, clip, width=0.975\textwidth]{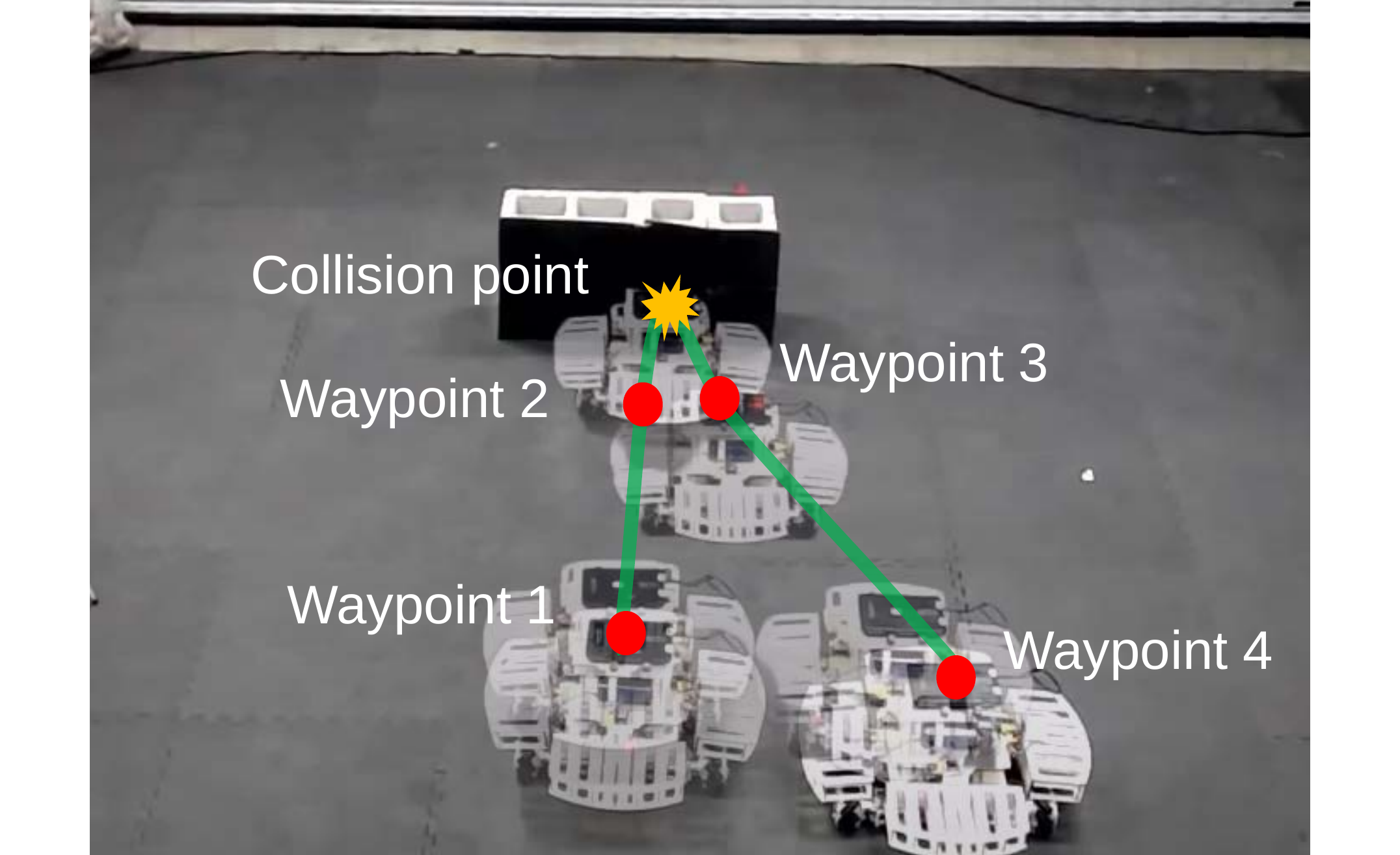}
      \end{subfigure}
      \begin{subfigure}{0.235\textwidth}
        \includegraphics[trim={6cm 0.2cm 6cm 5.5cm}, clip, width=0.975\textwidth]{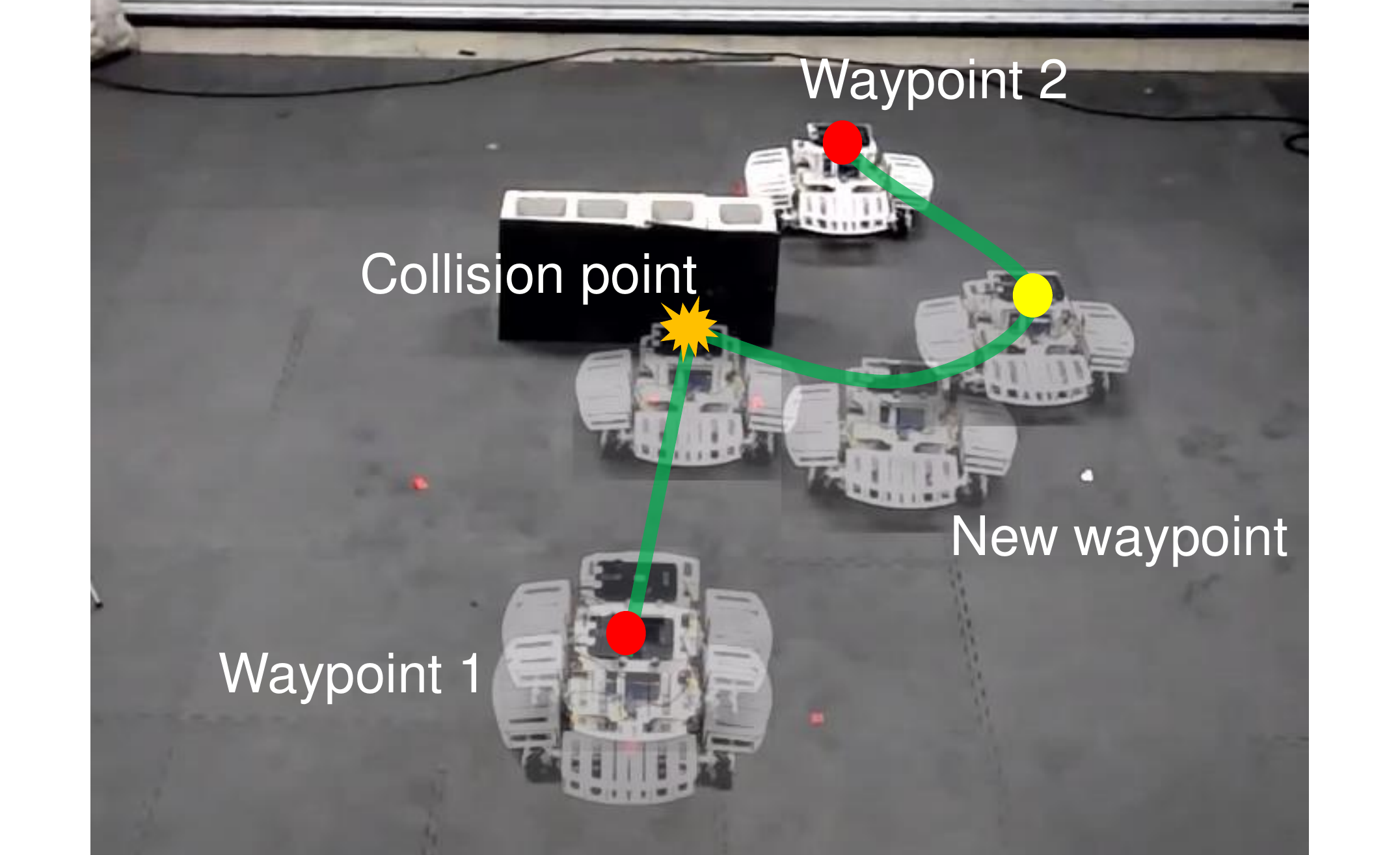}
      \end{subfigure}
      \caption{Composite images of a sample experiment with our proposed DRR strategy. The robot must go from start to goal passing through all waypoints, including intermediate ones created post-collision. Snapshots shown every $2$\;s. }
      \label{fig:ExperimentDRR}
	\vspace{-6pt}
\end{figure}

\begin{figure}[h!]
\vspace{-2pt}
 \begin{subfigure}{.235\textwidth}
  \centering
  \includegraphics[trim={4cm 0.2cm 4cm 0.2cm}, clip, width=0.9\linewidth]{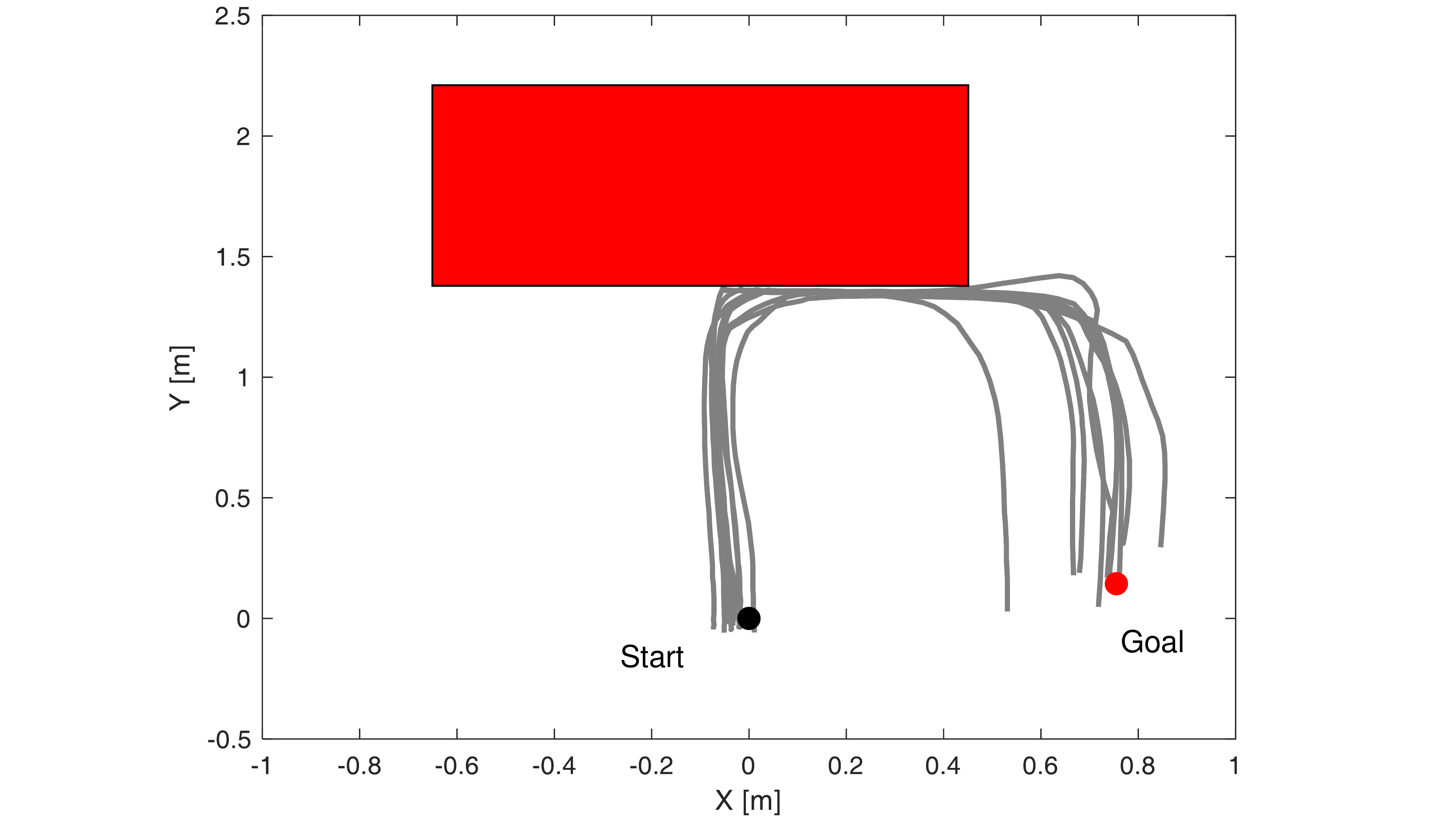}
  \vspace{-3pt}
  \caption{Case 1 collision avoidance}
 \end{subfigure}%
 \begin{subfigure}{.235\textwidth}
  \centering
  \includegraphics[trim={4cm 0.2cm 4cm 0.2cm}, clip,width=0.9\linewidth]{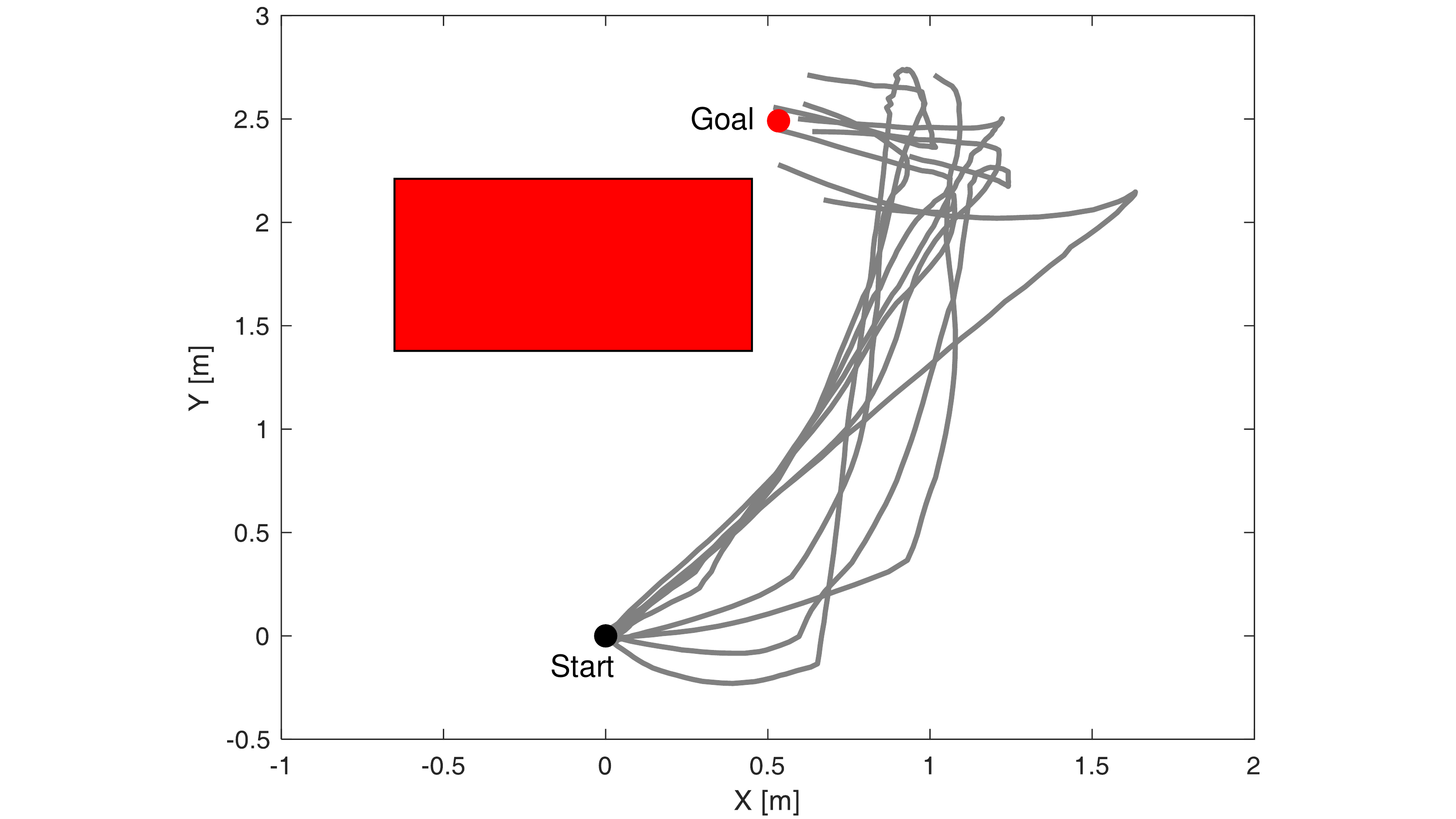}
  \vspace{-3pt}
  \caption{Case 2 collision avoidance}
 \end{subfigure}
 \vspace{-1pt}
 \medskip
 \begin{subfigure}{.235\textwidth}
  \centering
  \includegraphics[trim={4cm 0.2cm 4cm 0.2cm}, clip,width=0.9\linewidth]{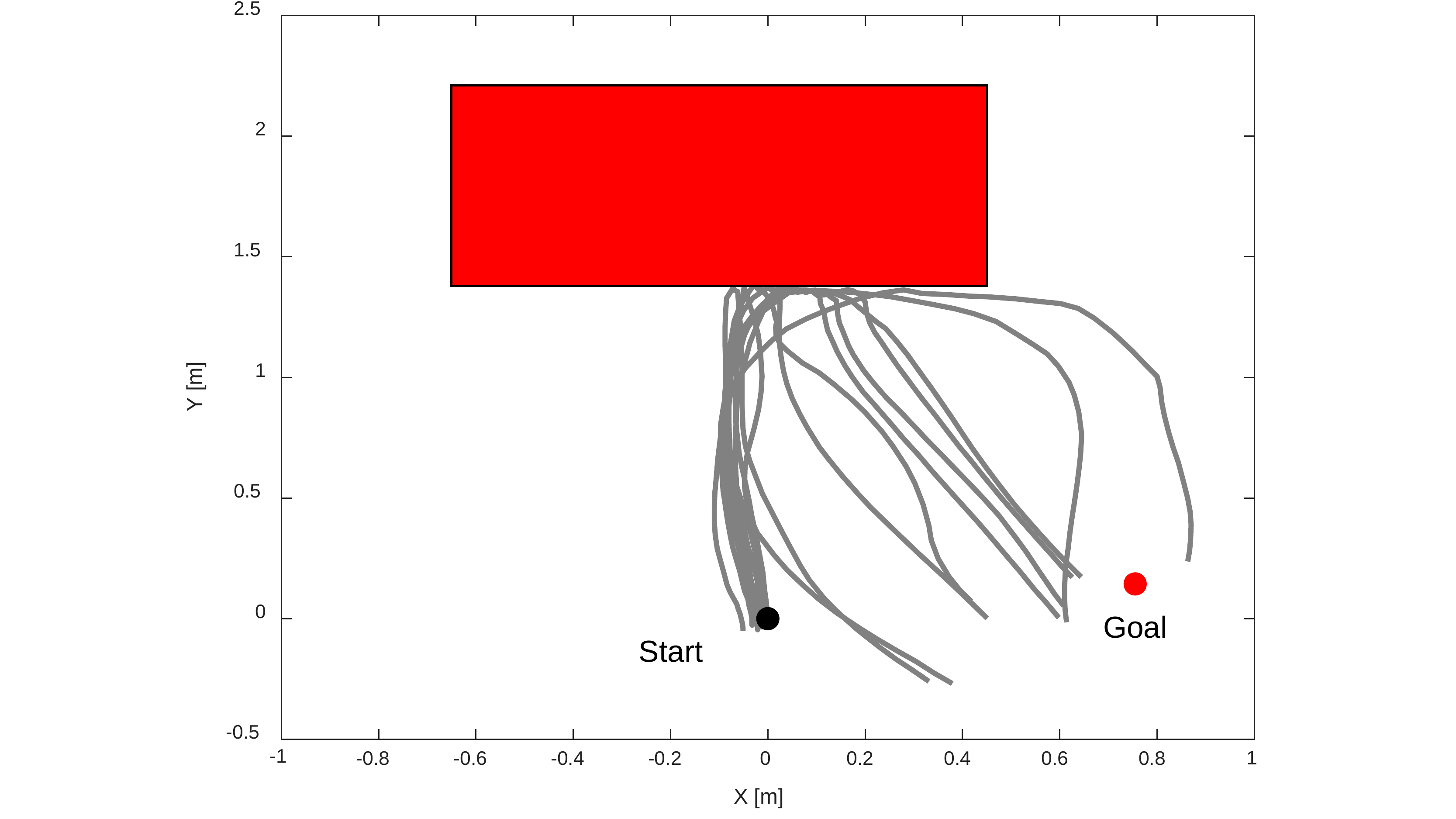}
  \vspace{-3pt}
  \caption{Case 1 DRR}
 \end{subfigure}%
 \begin{subfigure}{.235\textwidth}
  \centering
  \includegraphics[trim={4cm 0.2cm 4cm 0.2cm}, clip,width=0.9\linewidth]{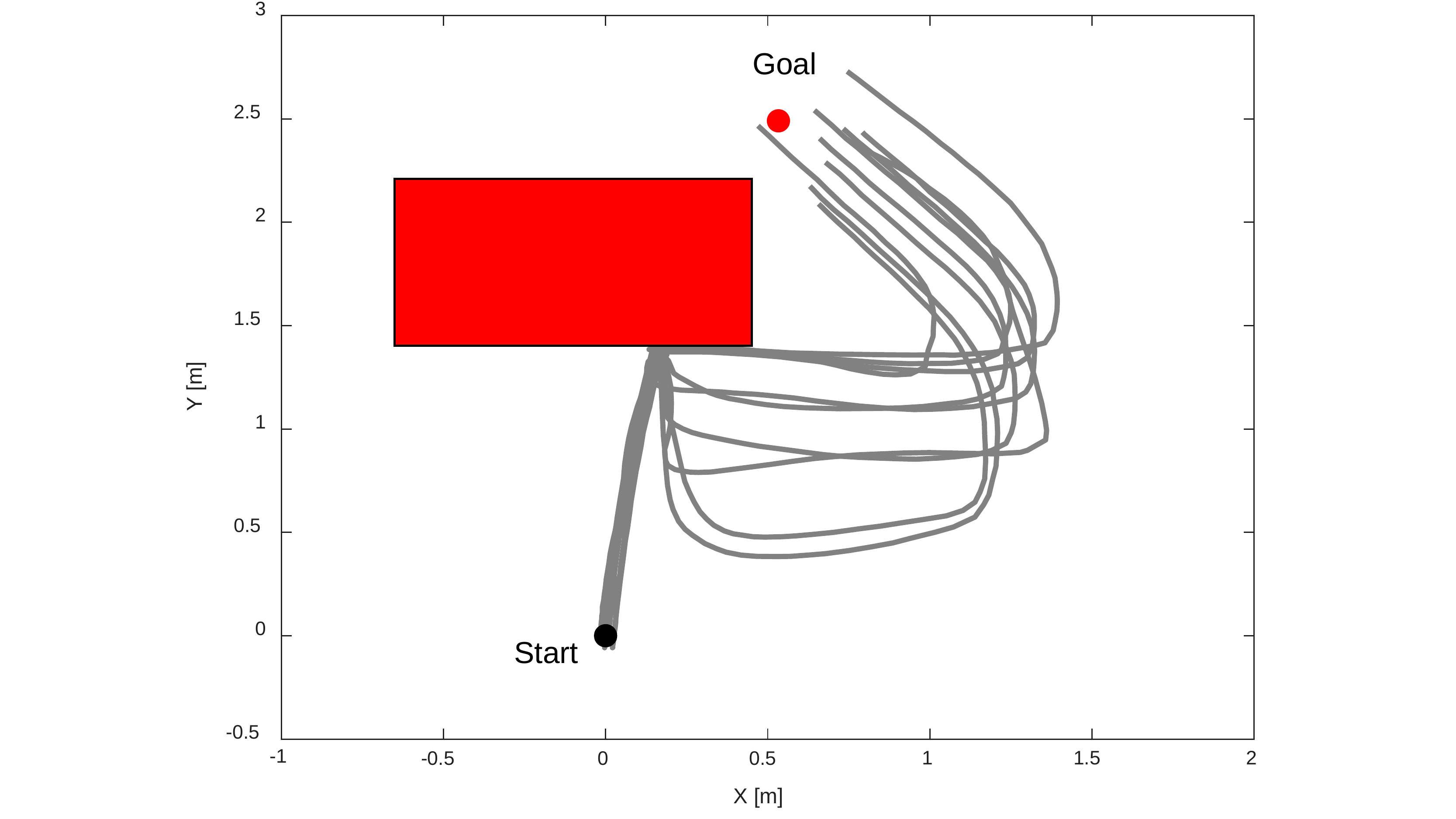}
  \vspace{-3pt}
  \caption{Case 2 DRR}
 \end{subfigure}%
 \vspace{-6pt}
 \caption{Experimental trajectories generated from DRR and collision avoidance trajectory generation strategy in \cite{richter2016polynomial} when the preplanned path intersects or does not intersect with the obstacle. (In all cases we conduct 10 trials). }
 \label{fig:compare DRR and avoidance}
 \vspace{-18pt}
\end{figure}

Case 1 tests the condition $~^{c}\bm{p}_{next, x} \geq -\rho$, i.e. no waypoint is added as per Alg.~\ref{pseudo replanner}. Case 2 tests $~^{c}\bm{p}_{next, x} < -\rho$, i.e. a waypoint is added to the list. 
In case 2, we run RRT* to generate a collision free path and perform path simplification to remove nodes without affecting the path's collision safety (Fig.~\ref{fig:path_in_case_2}).  The path simplification technique removes intermediate waypoints between two waypoints if a line segment between those two does not intersect with the obstacle.  
Then use the trajectory generation strategy in~\cite{richter2016polynomial}. We perform $10$ trials for each case. Instances of DRR and all experimental trajectories are shown in Fig.~\ref{fig:ExperimentDRR} and Fig.~\ref{fig:compare DRR and avoidance}.

Even though we design a collision-free desired trajectory with the strategy in \cite{richter2016polynomial}, the robot may still collide with the environment given for instance unmodeled dynamics such as drift. In case 1 there are $3$ out of $10$ trials that the robot in fact collides with the obstacle applying trajectory generation~\cite{richter2016polynomial} that aims to avoid collisions.  Table~\ref{table:compare} shows statistics on mean arrival times, path lengths and control energy.


\begin{table}[!h]
\vspace{-0pt}
    \caption{Comparison of trajectory generation strategy in \cite{richter2016polynomial} (Collision-avoidance) and DRR (Collision-inclusive) strategies.}
    \vspace{-6pt}
    \label{table:compare}
    \begin{center}
    \begin{tabular}{c c c c c}
    \toprule
    & \multicolumn{2}{c}{Strategy in \cite{richter2016polynomial}} & \multicolumn{2}{c}{DRR (our method)}\\
    \toprule
    & Case 1 & Case 2 & Case 1 & Case 2\\
    \toprule
    $\bar{T}_{end}$ [$s$] 
    & $7.7$
    & $8.71$
    & $6.16$
    & $9.17$\\
    \midrule
    $STD(T_{end})$ 
    & $0$
    & $2.20$
    & $0.22$
    & $0.31$\\
    \midrule
    $\bar{s}$ [$m$]
    & $3.153$
    & $3.448$
    & $2.977$
    & $4.38$\\
    \midrule
    $STD(s)$ 
    & $0.117$
    & $0.495$
    & $0.150$
    & $0.451$\\
    \midrule
    $\bar{E_{c}}$ [$m^{2}/s^{3}$]
    & $56.83$
    & $80.62$
    & $58.42$
    & $255.96$\\
    \midrule
    $STD(E_{c})$ 
    & $29.34$
    & $54.38$
    & $32.92$
    & $133.54$\\
    \bottomrule
    \end{tabular}
    \end{center}
    \vspace{-18pt}
\end{table}

In case $1$ for DRR, mean arrival times $\bar{T}_{end}$ and path lengths $\bar{s}$ decrease by $25\%$ and $6\%$, while the control energy increases by $2.8\%$ on average. However, the error in the end point increases by $25\%$. In case $2$, mean arrival times and path lengths increase by $5.2\%$ and $27\%$, and control energy increases by $258\%$. 
This is because the output velocity of DRR is not flat since the robot needs to decelerate and then accelerate during the boundary following process. Moreover, the path generated by the boundary following is not the shortest.
However, since the path between the collision point and the new inserted waypoint is close to the obstacle surface, the existence of the obstacle decreases the control error in free space. The error in the end point decreases by $12\%$. Overall, these results show the tradeoff between online reactive execution (whereby collision checking is skipped) and collision avoidance.  

\section{Conclusions}

\vspace{-1pt}
The paper contributes to collision-inclusive motion planning, and proposes a new deformation recovery and replanning strategy to generate local reactive replanned trajectories following a collision sensed by the robot at run-time. Two novel components make this strategy possible: 1) a deformation recovery controller that optimizes robot states during post-impact recovery,
and 2) a post-impact trajectory replanner that adjusts the next waypoint with the information from the collision for the robot to pass through and generates a polynomial-based trajectory. Our proposed strategy runs online, and, given a sequence of waypoints that can be obtained in any manner, enables algorithmic collision resilience in a blind navigation paradigm. 

Comparisons with a collision-avoidance trajectory generation method reveal fundamental tradeoffs between collision-avoidance and collision-inclusive motion planning and control. A key insight is that a collision-inclusive strategy may apply better when abrupt changes in robot motion are required (as in Case 1 testing), whereby collisions can be exploited to reduce mean arrival times and path lengths. We anticipate that exploiting collisions as described in this work can complement sensor-based autonomous navigation in cluttered environments where finding obstacle-free paths may be too computationally expensive.
\vspace{-3pt}

\end{document}